\pdfoutput=1

\documentclass[11pt]{article}

\usepackage[]{acl}

\usepackage{times}
\usepackage{latexsym}

\usepackage[T1]{fontenc}

\usepackage[utf8]{inputenc}

\usepackage{microtype}

\usepackage{amsmath}
\usepackage{enumitem}
\usepackage{float}
\usepackage{graphicx}
\usepackage{multirow}

\usepackage[linesnumbered, ruled]{algorithm2e}
\usepackage[T1]{fontenc}
\usepackage{dsfont}

\usepackage{mdframed}
\usepackage{booktabs}
\usepackage{graphicx}
\usepackage{subcaption}

\usepackage[normalem]{ulem}
\usepackage{enumitem}
\usepackage{xcolor}
\usepackage{soul}
\colorlet{soulred}{red!30}
\sethlcolor{soulred}%

\usepackage{amsmath,amsthm,amsfonts,amssymb,bm}
\usepackage{framed}
\usepackage{varioref}
\usepackage{float}
\usepackage{multirow}
\usepackage{dashrule}
\usepackage{url}
\usepackage{pifont}
\usepackage{helvet}

\newcommand{\ie}{\emph{i.e., }}

\definecolor{green}{rgb}{0.1,0.1,0.1}
\definecolor{chocolate}{HTML}{D2691E}
\definecolor{maroon}{HTML}{A00000}
\definecolor{indigo}{HTML}{4B0082}
\definecolor{green}{HTML}{008000}
\definecolor{red}{HTML}{a91e1e}
\definecolor{cadmiumgreen}{rgb}{0.0, 0.42, 0.24}

\usepackage{bbding}
 
\usepackage{amssymb}
\usepackage{pifont}

\makeatletter
\newcommand*\myfontsize{%
  \@setfontsize\myfontsize{8}{9}%
}
\makeatother

\makeatletter
\newcommand*\mysmallfontsize{%
  \@setfontsize\mysmallfontsize{7.4}{8.4}%
}
\makeatother

\usepackage{adjustbox}

\newcommand{\myskip}[1]{}

\title{
A Comprehensive Analysis for Visual Object Hallucination \\in Large Vision-Language Models
}

\author{Liqiang Jing$^1$\thanks{\quad Equal contribution}, {\bf  Guiming Hardy Chen}$^1$\footnotemark[1], {\bf  Ehsan Aghazadeh}$^2$, {\bf Xin Eric Wang}$^3$, {\bf Xinya Du}$^1$ \\   $^1$University of Texas at Dallas, $^2$University of Massachusetts at Amherst, $^3$University of California, Santa Cruz \\ jingliqiang6@gmail.com, \{guiming.chen,xinya.du\}@utdallas.edu}

\begin{document}
\maketitle


\begin{abstract}
Large Vision-Language Models (LVLMs) demonstrate remarkable capabilities in multimodal tasks, but visual object hallucination remains a persistent issue. It refers to scenarios where models generate inaccurate visual object-related information based on the query input, potentially leading to misinformation and concerns about safety and reliability. Previous works focus on the evaluation and mitigation of visual hallucinations, but the underlying causes have not been comprehensively investigated. In this paper, we analyze each component of LLaVA-like LVLMs—the large language model, the vision backbone, and the projector, to identify potential sources of error and their impact. Based on our observations, we propose methods to mitigate hallucination for each problematic component. Additionally, we developed two hallucination benchmarks: QA-VisualGenome, which emphasizes attribute and relation hallucinations, and QA-FB15k, which focuses on cognition-based hallucinations. 
\end{abstract}

\section{Introduction}
\label{sec:intro}


Large Language Models (LLMs), such as GPT-3 \cite{gpt3} and ChatGPT \cite{chatgpt}, have showcased remarkable proficiency in language tasks, yet they encounter significant challenges when it comes to processing multimodal inputs. This limitation has driven a shift in research towards Large Vision-Language Models (LVLMs)~\cite{llava,mPLUG-Owl,llava_rlhf}, which integrate advanced LLMs~\cite{llama,vicuna} with Vision Foundation Models (VFMs)~\cite{vit,DBLP:journals/corr/abs-2108-07258} to enhance multimodal understanding. LVLMs have demonstrated impressive capabilities across various tasks that require visual and textual integration, including Visual Question Answering~\cite{vqatask}, Image Captioning~\cite{mscoco}, and Visual Entailment~\cite{DBLP:conf/aaai/ZhangJG25}.

\begin{figure}
    \centering
\includegraphics[width=\linewidth]{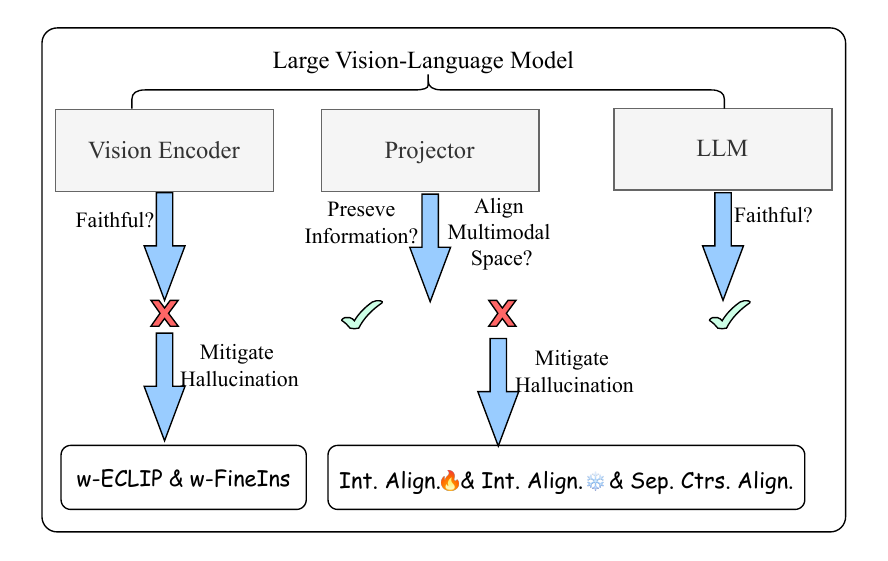}
    \caption{
    An overview of our paper. We first investigate the sources of hallucination from a component-level perspective within the LVLM architecture. Based on the identified causes, we then design targeted methods to mitigate hallucinations effectively.
    }
    \label{fig:intro}
\end{figure}

Despite these advances, visual hallucination remains a persistent issue in LVLMs ~\cite{DBLP:conf/emnlp/RohrbachHBDS18,liu2023aligning,liu2023hallusionbench,DBLP:journals/corr/abs-2310-16045,zhang2024fine}. This phenomenon occurs when models generate inaccurate or misleading information unrelated to the actual visual input, potentially leading to misinformation and raising concerns about safety and reliability in real-world applications ~\cite{DBLP:conf/emnlp/LiDZWZW23}.
Visual object hallucination, including object existence, attribute, and relation, has garnered significant attention due to its widespread occurrence in images. 
Current works on visual object hallucination mainly focus on evaluation and mitigation. For example, \citet{DBLP:conf/emnlp/LiDZWZW23} extends CHAIR \cite{DBLP:conf/emnlp/RohrbachHBDS18} and proposes POPE, a polling-based query technique for probing object-level hallucination. For hallucination mitigation, \citet{2023llavarlhf} introduce new alignment algorithm called Factually Augmented RLHF that augments the reward model with additional factual information such as image captions and ground-truth multi-choice options, which alleviates the reward hacking phenomenon in RLHF and further improves the performance.

\textcolor{black}{While existing works have achieved notable success in visual object hallucination, they lack a comprehensive component-level analysis of the model architecture to pinpoint where and how hallucinations occur.}
In this work, we focus on visual object-related hallucination and LLaVA-like LVLMs, which typically consist of three modules: the large language model (LLM), the vision backbone, and the projector.
Errors in any of these modules can lead to issues in the overall performance or functionality of the model. Therefore, we conduct an independent analysis of each component to identify potential sources of error and their impact. From our study, we have the following findings. 
1) The LLM in LVLM is able to generate faithful content when captions of images are provided as input.
2)	Hallucinations exist in the perception process of the vision backbone.
3)	Projector is able to preserve visual features, but has trouble aligning between visual and textual spaces.


Based on our observations, we propose methods for the two problematic components to mitigate their hallucination issue.
To improve the \textbf{vision backbone}, we propose to finetune CLIP with fine-grained data and fine-grained perception-based visual instruction tuning, and find that both of them can reduce hallucination caused by the vision backbone.
For the \textbf{projector}, we propose a contrastive alignment objective with three variations, which can all be integrated into the original training pipeline with minimal additional costs.


To conduct a comprehensive hallucination evaluation, we develop a fine-grained hallucination benchmark named QA-VisualGenome, which is built upon the Visual Genome dataset~\cite{DBLP:journals/ijcv/KrishnaZGJHKCKL17}. Unlike existing object-oriented hallucination benchmarks (\textit{e.g.}, POPE), QA-VisualGenome emphasizes the detailed attribute and relationship hallucinations.
Furthermore, existing hallucination benchmarks primarily focus on perception-based hallucinations for general objects, neglecting cognition-based hallucinations such as the names of people and famous buildings. To address this gap, we construct a cognition-based hallucination benchmark named QA-FB15K, which is based on the FB-15K dataset \cite{DBLP:conf/nips/BordesUGWY13}, a multimodal knowledge graph with textual entities, image entities, and textual relations. QA-FB15K presents challenges for models in leveraging world knowledge to solve the questions. 

Our main content is shown in Figure \ref{fig:intro}. Our contributions can be summarized as follows:
1) We analyze the hallucination caused by each component in LVLMs and provide component-wise takeaway messages.
2) Based on our observation, we propose several methods to improve each hallucinated component.
3) We construct a fine-grained hallucination benchmark based on Visual Genome and a cognition-based hallucination benchmark based on FB15k for evaluation.
4) We extensively evaluate our proposed methods on various benchmarks, and provide in-depth analysis\footnote{All benchmark datasets, code, and models will be released.}. 


\section{Hallucination Analysis}
LVLMs consist of three components:  language decoder $\mathcal{D}$, projector vision encoder $\mathcal{V}$, and $\mathcal{P}$.
We first introduce the datasets for evaluation and then provide in-depth analysis for each component.

\subsection{Settings}
\label{sec:analysis_settings}

We select two benchmarks to benchmark the performance of each component. 
1) \textbf{POPE} \cite{DBLP:conf/emnlp/LiDZWZW23}.  
POPE is a benchmark designed for evaluating object existence hallucinations in LVLMs, incorporating three sampling methods for generating negative samples: random, popular, and adversarial. 
In the random setting, objects not present in the image are randomly selected. 
In the popular setting, negative samples are drawn from a pool of frequently occurring objects. 
In the adversarial setting, the sampling focuses on objects that frequently co-occur with present objects but do not actually exist in the image. 
2) \textbf{QA-VisualGenome}. To further investigate the hallucination issue on relations and attributes of objects, we construct a new fine-grained evaluation benchmark based on the VisualGenome dataset \cite{DBLP:journals/ijcv/KrishnaZGJHKCKL17}, which collects dense annotations of attributes and relationships of objects for each image. 
Specifically, we design two types of Yes-or-No questions to evaluate models: attributes and relations. 
For example, an attribute question could be ``\textit{Is the dog red in the image?}'' 
A relational question would ask, ``\textit{Is the dog standing on the table?}''. 
Similar to previous work \cite{DBLP:conf/cvpr/WangHZS20a}, we exclude uncommon relations and attributes. We randomly select one relation or attribute to generate negative samples.


\begin{table*}[h!]
\centering
\vspace{-20pt}
\caption{Performance (\%) of LLMs across different datasets when visual information is provided in textual format.
\textit{LLaVA}: image+text query as input on original LLaVA model; 
\textit{Vicuna}: caption+text query as input on Vicuna-1.5;
\textit{Vicuna$_{\text{LLaVA}}$}: caption+text query as input on the Vicuna model in LLaVA (LLM undergone visual instruction tuning).
}
\resizebox{.8\linewidth}{!}{
\begin{tabular}{l|cc|cc|cc|cc|cc}
\toprule
\multirow{3}{*}{\textbf{Model}} & \multicolumn{6}{c|}{\textbf{POPE}} & \multicolumn{4}{c}{\textbf{QA-VisualGenome}} \\ 
\cmidrule(lr){2-7} \cmidrule(lr){8-11} 
 & \multicolumn{2}{c|}{\textbf{Random}} & \multicolumn{2}{c|}{\textbf{Popular}} & \multicolumn{2}{c|}{\textbf{Adversarial}}  & \multicolumn{2}{c|}{\textbf{Attribute}} & \multicolumn{2}{c}{\textbf{Relation}} \\ 
\cmidrule(lr){2-11}
 & \textbf{Acc} & \textbf{F1} & \textbf{Acc} & \textbf{F1} & \textbf{Acc} & \textbf{F1} & \textbf{Acc} & \textbf{F1} & \textbf{Acc} & \textbf{F1}  \\ 
\midrule
\textit{LLaVA-7B} & 87.42 & 86.36 & 86.63 & 85.25 & 85.13 & 83.88&64.67 &66.60& 67.57& 74.81 \\ 

\textit{Vicuna-7B} & 92.67&92.09&92.67&92.09&93.00&92.47 & 57.23 & 69.83 & 79.50&\textbf{80.79} \\
\textit{Vicuna-7B$_{\text{LLaVA}}$} & 100.00 & \textbf{100.00} & 100.00 & \textbf{100.00} & 99.67 & \textbf{99.67} &68.29&\textbf{75.92}&63.2&{73.06}\\
\midrule
\textit{LLaVA-13B} & 91.33 &91.72 &88.33 & 89.16&84.33&85.97&55.99&68.86 &56.40 &69.38 \\ 
\textit{Vicuna-13B} & 87.90 & 89.15 & 95.00 & 95.24  &  90.00&90.91  & 87.90&\textbf{89.15} & 87.90 & \textbf{89.25} \\ 
\textit{Vicuna-13B$_{\text{LLaVA}}$ }& 99.67 & \textbf{99.67} & 99.67 & \textbf{99.60} & 99.33 & \textbf{99.33}  & 75.41 & 80.10 & 84.30 & 84.29 \\ 
\bottomrule
\end{tabular}
\label{tab:llm_analysis}
}
\end{table*}

\subsection{Language Decoder}
\label{sec:llm_probing}

\paragraph{Conjecture 1. LLM in LVLM is able to generate faithful content when image captions are provided as input. }

To validate this conjecture, we use the POPE dataset to evaluate the performance of LLMs. Instead of providing images to the LVLMs, we only input text descriptions of the images. 
For POPE, we obtain objects from the MSCOCO~\citep{mscoco} dataset and feed the LVLM with objects in the image and the textual query from POPE to generate the response.
For QA-VisualGenome, we feed the LVLM with objects, object attributes, and relations presented in the image to replace visual information. 
This helped assess the model's ability to hallucinate when provided with accurate textual descriptions of the image.  
In addition, we also test the original Vicuna as a baseline.

We show the performance of LLMs in Table \ref{tab:llm_analysis}.
From the results, we found that the performance will be improved largely if we provide the correct visual information in a textual format. This indicates the current main reason for hallucination is caused by a vision encoder or projector. Specifically, 
the model could achieve an accuracy of 99.67\% when provided with complete object descriptions for the random setting of POPE, which shows the LLM is robust when given the correct information about the whole image. 
In addition, we also found that the LLM after the pertaining and instruction tuning of LLaVA performs better than the original LLM. LLaVA fine-tuning likely enhances the model's object recognition, memory of object-specific features, instruction-following ability, and contextual understanding of visual descriptions, enabling it to accurately identify common objects within text descriptions even without actual images.


\begin{table}[t]
\centering
\caption{Performance of CLIP in the text-image matching across different datasets measured by Accuracy (\%).}
\label{tab:clipresults}
\resizebox{.85\columnwidth}{!}{
\begin{tabular}{ccc|cc}
\toprule
 \multicolumn{3}{c|}{\textbf{POPE}} & \multicolumn{2}{c}{\textbf{QA-VisualGenome}} \\  \cmidrule(lr){1-5}
 Random & Popular & Adversarial & Attribute & Relation \\ \midrule
 83.33 & 87.30 & 86.00 & 61.57 & 60.22 \\ 
\bottomrule
\end{tabular}}
\end{table}


\subsection{Vision Encoder}
\paragraph{Conjecture 2. There are hallucinations in the perception process of the vision encoder. } 
To verify this factor, we conducted experiments using CLIP on a text-image matching task. Specifically, we designed a template of the form "There is a/an \{object\} in the image," where \{object\} corresponds to various objects in the input images. 
For each image, we assigned one ground-truth object and a hallucinated object for the template.  We use accuracy as the evaluation metric.
We show all the experimental results in Table \ref{tab:clipresults}.
Overall, we found that the performance of CLIP on the text-matching task is not good. 
For example, the performance of CLIP on the text-image matching task is 83.33\% accuracy on the random setting of POPE, indicating the presence of hallucinations within the vision encoder's perception process. 

Another interesting phenomenon is that the accuracy of CLIP in recognizing objects is worse than LLaVA, even the LLaVA adopts CLIP as the vision encoder. Specifically, the accuracy of LLaVA is 91.33\% on the random setting of POPE, but CLIP only achieves 83.33\% accuracy. This indicates that the hallucination caused by CLIP can be alleviated to a certain extent after the pre-training feature alignment and instruction tuning.
The potential reason may be that  LLaVA’s training uses diverse questions aligned with specific image features, optimizing for generative loss. This fine-grained alignment helps the model better understand and describe visual content with greater accuracy and detail.

\subsection{Projector}
\label{sec:analysis_projector}
We analyze the projector module from two perspectives corresponding to its two roles in the LVLM: \textit{preserving visual information} and \textit{aligning visual and textual spaces}.

\textbf{Conjecture 3. The projector should not result in significant visual information loss.}
We formalize the hypothesis using the notion of \textit{V-information}~\citep{hewitt2021conditional}. Let \( \Phi_{\textit{pre}}(X) \) and \( \Phi_{\textit{post}}(X) \) represent the pre-projector and post-projector representations, respectively. We compare the V-information between these representations and a target property \( Y \) (e.g., a classification label).

We define the V-information for pre- and post-projector representations as
\[
I_{\mathcal{V}}(\Phi_{\textit{pre}}(X) \rightarrow Y) = H_{\mathcal{V}}(Y) - H_{\mathcal{V}}(Y | \Phi_{\textit{pre}}(X))
\]
\[
I_{\mathcal{V}}(\Phi_{\textit{post}}(X) \rightarrow Y) = H_{\mathcal{V}}(Y) - H_{\mathcal{V}}(Y | \Phi_{\textit{post}}(X))
\]
where \( H_{\mathcal{V}} \) is the \textit{V-entropy}~\citep{hewitt2021conditional}. \( H_{\mathcal{V}}(Y) \) is the entropy of \( Y \), which reflects the inherent uncertainty of \( Y \) without any conditioning on the representations. \( H_{\mathcal{V}}(Y | \Phi(X)) \) represents the uncertainty we have in predicting \( Y \) after observing the representation \( \Phi(X) \), using functions from the family \( \mathcal{V} \). It is formally defined as:
\[
H_{\mathcal{V}}(Y | \Phi(X)) = \inf_{f \in \mathcal{V}} \mathbb{E}_{\Phi(X), Y} \left[ -\log f(\Phi(X))(Y) \right]
\]
This expression measures the best performance that a function \( f \) from the function family \( \mathcal{V} \) can achieve when predicting \( Y \) given the representation \( \Phi(X) \). The lower this value, the more predictive power the representation \( \Phi(X) \) has regarding \( Y \).

The goal is to determine whether information loss occurs in the projection layer. If the projection layer introduces no information loss, then the V-information of the pre-projector and post-projector representations should be approximately equal:
\[
I_{\mathcal{V}}(\Phi_{\textit{pre}}(X) \rightarrow Y) = I_{\mathcal{V}}(\Phi_{\textit{post}}(X) \rightarrow Y)
\]

We compare the V-information accessible from both the pre-projector and post-projector representations. The performance of a probe (e.g., classifier) trained on \( \Phi_{\textit{pre}}(X) \) and \( \Phi_{\text{post}}(X) \) provides an empirical estimate of these quantities:
\[
\textit{Perf}_{\textit{pre}} = \max_{\theta} \mathbb{E}[ \log P(Y | f_{\theta}^{\textit{pre}}(\Phi_{\textit{pre}}(X)))]
\]
\[
\textit{Perf}_{\textit{post}} = \max_{\theta} \mathbb{E}[ \log P(Y | f_{\theta}^{\textit{post}}(\Phi_{\textit{post}}(X)))]
\]

To determine if information loss occurs, we compute the difference in performance:
\[
\Delta \textit{Perf} = \textit{Perf}_{\textit{pre}} - \textit{Perf}_{\textit{post}}
\]

If \( \Delta \textit{Perf} = 0 \), this implies that no information loss has occurred and the information available in \( \Phi_{\textit{pre}}(X) \) is fully retained in \( \Phi_{\textit{post}}(X) \). However, if \( \Delta \textit{Perf} > 0 \), this indicates that the post-projector representation has lost some information present in the pre-projector representation, leading to a decrease in predictive power for \( Y \).

With the hypothesis grounded to V-information, we conduct a probing experiment on LLaVA-7B to verify it. We linear-probe the pre- and post-projector feature with image classification tasks on CIFAR10~\citep{krizhevsky2009learning}, CIFAR100~\citep{cifar100} and ImageNet~\citep{deng2009imagenet}. 
Results in Table~\ref{tab:projector_cls_probing} shows that for the 13B LLaVA model, performance percentage drop of post-projection features is less than 2\%, indicating that the visual features are well preserved by the projectors in both models.

\begin{table}[h]
\centering
\caption{Performance of linear probing using pre- and post-projector image features on CIFAR10, CIFAR100 and ImageNet. Accuracy\% is used as the metric.}
\resizebox{0.6\columnwidth}{!}{
\begin{tabular}{l|cl}
\toprule
\multirow{2}{*}{Dataset}  & \multicolumn{2}{c}{\textit{LLaVA-13B}} \\
& $\textit{Perf}_\textit{pre}$ & $\textit{Perf}_\textit{post}$  \\ \hline
CIFAR10  & 96.27 & $96.15_\text{ \color{red}-0.12\% }$ \\ 
CIFAR100 &  81.78 & $81.02_\text{ \color{red}-0.93\% }$ \\
ImageNet & 71.97 & $70.83_\text{ \color{red}-1.58\% }$ \\ 
\bottomrule
\end{tabular}
}
\label{tab:projector_cls_probing}
\end{table}


\textbf{Conjecture 4. The projector should align the visual and textual spaces.}
As its name suggests, the projector should be able to project the source (visual) space to the target (textual) space. 
To probe the alignment between two spaces, we collect caption data from MSCOCO~\citep{mscoco}, LLaVA-Caption~\citep{llava15}, ALLaVA~\citep{chen2024allava} and compute the similarity between a projected image feature and the textual embedding of its caption. 
The rationale of using cosine similarity is that, based on the findings in Section~\ref{sec:llm_probing}, a large performance boost is observed if we replace an image with its caption. Therefore, if the projected image feature is similar enough to its caption embedding (\textit{i.e.} cosine similarity=1), then an LVLM should gain similar performance to the case where an image is replaced by its caption as input.

Results in Table~\ref{tab:projector_emb_probing} show that the cosine similarities of the two features are fairly low, indicating nearly independent relationships. This finding is consistent with the existing work \citep{huang2024deciphering,DBLP:conf/aaai/LiGZCSMK25}, which reveals that visual and textual representations are apart from each other in the embedding space. 
Therefore, the projector in LLaVA models may not function as an alignment module as well as expected, which could be one of the causes of hallucination for the entire model.

\begin{table}[]
\centering
\caption{Cosine similarity between projected image features and textual embedding of corresponding captions across different datasets. Captions are processed by Vicuna~\citep{vicuna} tokenizer.}
\resizebox{\columnwidth}{!}{
\begin{tabular}{l|cccc}
\toprule
\multirow{2}{*}{Dataset} & \multirow{2}{*}{Token Length} & \multirow{2}{*}{Image Res.} & \multicolumn{2}{c}{Cos. Sim.} \\
 & & & 7B & 13B \\ \hline
 MSCOCO & 15.16 & (575, 488)  & 0.03 &  0.04 \\
 LLaVA Caption & 15.09 & (412, 366) & 0.03 & 0.04 \\
 ALLaVA & 222.83 & (1020, 923) & 0.05 & 0.06 \\
\bottomrule
\end{tabular}
}

\label{tab:projector_emb_probing}
\end{table}









\section{Mitigating Object Hallucination Caused by  Different Modules}
Based on the analysis in Section 2, we further devised different methods to mitigate the object hallucination in different components in LVLMs.

\subsection{How to alleviate the hallucination caused by CLIP?}

\begin{figure}
    \centering
\includegraphics[width=\linewidth]{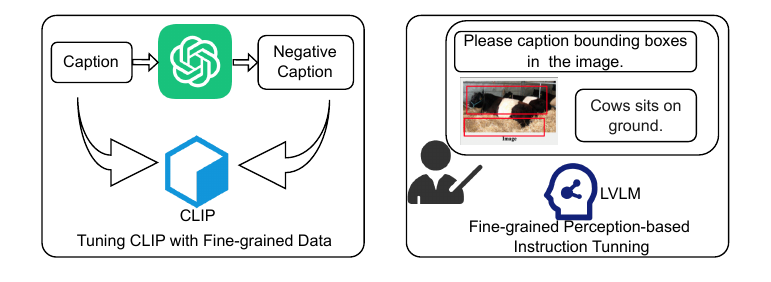}
    \caption{Tuning CLIP with fine-grained data (left) and fine-grained perception-based instruction tuning (right).}
    \label{fig:method1}
\vspace{-10pt}
\end{figure}

As previously noted, the vision backbone within LVLMs also contributes to hallucinations. The CLIP model, as the vision encoder of LLaVA, is trained on massive image-caption pairs from the internet with a contrastive loss objective. 
However, these captions are typically brief and noisy,
and negative pairs often differ substantially from positive ones.
Therefore, it is likely that the model can distinguish them without needing to capture the finer details in the images. Consequently, the model may achieve high accuracy while lacking a nuanced understanding of the visual content \cite{DBLP:conf/emnlp/LiuJSWZ24}.
To address this issue, we propose two methods to reduce hallucination caused by the vision backbone, as shown in Figure \ref{fig:method1}.

\textbf{Tuning CLIP with fine-grained data}
A direct method to improve CLIP is to post-train CLIP with more fine-grained samples. This is because the CLIP is trained with massive images paired with brief captions.
In this method, we leverage GPT-4 \cite{chatgpt} to generate negative examples, which are then used in a contrastive learning setup to improve the discriminative ability of CLIP.

\textit{Generate Negative Examples:}  Inspired by prior work indicating that LVLMs are more likely to generate hallucinatory responses for frequently occurring objects \cite{DBLP:conf/emnlp/LiuJSWZ24}, we devise two strategies: inserting hallucinatory objects and removing existing ones. 

For the insertion strategy, we categorize objects in images into three types—random, popular, and adversarial—each containing three objects. Random objects are sampled randomly, popular objects are the top frequent objects in the whole dataset, and adversarial objects are the top frequent objects with the current objects. 
By inserting one to three objects from each category into the correct captions with the assistance of GPT-4, we create examples with varying levels of hallucinations (\ie negative samples). 
For the removal strategy, we randomly select one or two segmented objects from the caption and instruct GPT-4 to eliminate them from the caption. 


\textit{Contrastive Learning}: 
We use these generated negative examples in a contrastive learning framework where CLIP is trained to correctly distinguish between the positive and negative pairs. By exposing the model to these fine-grained differences, CLIP becomes better at understanding nuanced visual features.

First, let \(I\) represent an image embedding and \(T\) a text embedding. Let \(T^{+}\) be the text vector that correctly matches \(I\), and let \(T^{-}\) denote a collection of negative texts not semantically aligned with \(I\). We also introduce \(\beta\) as a temperature parameter. 

The fundamental image-to-text contrastive objective can be expressed as:
\begin{equation}
\mathcal{L}_{i2t} \;=\; -\,\log ( \frac{\exp({I \cdot T^{+}}/{\beta})}{\sum_{T^{*} \in \{T^{+}, T^{-}\}}\exp({I \cdot T^{*}}/{\beta})} ).
\end{equation}
The symmetric term $\mathcal{L}_{t2i}$ can be constructed for text-to-image alignment.
Combining them yields the image-text contrastive loss:
\begin{equation}
\label{eq:contrastive_loss}
\mathcal{L}_{itc} \;=\; \tfrac{1}{2}(\mathcal{L}_{i2t} \,+\, \mathcal{L}_{t2i}).
\end{equation}

Next, consider that we introduce an additional set of artificially generated negative texts \(\{T^{neg}\}\). 
Incorporating these into the image-to-text objective gives:
\begin{equation}
    \mathcal{L}_{i2t} \;=\; -\,\log ( \frac{\exp({I \cdot T^{+}}/{\beta})}{\sum_{T^{*} \in \{T^{+}, T^{-}, T^{neg}\}}\exp({I \cdot T^{*}}/{\beta})} ).
\end{equation}

To further refine the separation between correct matches and all classes of negative samples (both standard and synthetic), we introduce a margin-based term. Let \(\tau_{1}\) be the margin threshold enforcing that a positive pair’s similarity should exceed that of any negative pair by at least \(\tau_{1}\):
\begin{equation}
    \mathcal{L}_{1} \;=\; \max\bigl(0,\; \tau_{1} - (I \cdot T^{+}) + (I \cdot T^{\star})\bigr),
\end{equation}
where \(T^{\star}= \{T^{-}, T^{neg}\}\) is the union of standard and synthetic negatives.

Additionally, to encourage the model to distinguish synthetic negatives from standard negatives—thus capturing subtle semantic cues—we introduce another margin loss. Let \(\tau_{2}\) control the required margin between these two types of negative samples:
\begin{equation}
\mathcal{L}_{2} \;=\; \max\bigl(0,\; \tau_{2} - (I \cdot T^{neg}) + (I \cdot T^{-})\bigr).    
\end{equation}

Finally, assigning weighting factors \(\lambda_{1}\) and \(\lambda_{2}\) to the margin terms allows adaptive emphasis on these constraints. The complete objective function is:
\begin{equation}
    \mathcal{L} \;=\; \mathcal{L}_{itc} \;+\; \lambda_{1}\mathcal{L}_{1} \;+\; \lambda_{2}\mathcal{L}_{2}.
\end{equation}
This integrated loss framework guides the model to better discriminate correct image-text pairs from both standard and refined negative samples.

\textbf{Fine-grained perception-based visual instruction tuning} 
As we mentioned, CLIP may not capture the finer details in the visual representation from the vision encoder. Therefore, we attempt to enable the LLM to perceive the fine-grained information within the CLIP vision encoder. Meanwhile, the method of enhancing CLIP and then replacing it is time-consuming, as it requires additional steps for feature alignment and instruction tuning after replacing the vision encoder of LVLMs. As a result, we explore a more efficient approach by directly enabling the LLM to perceive the detailed visual features during visual instruction tuning. 

To achieve this, we propose \textit{fine-grained perception-based visual instruction tuning}. Specifically, we randomly select two bounding boxes from the image, and then use the object attributes corresponding to these bounding boxes and their relationships to generate the corresponding captions. We then create instruction tuning data $(I_f, T_f, R_f)$, where $T_f$ is the textual prompt: ``Please caption the content in the bounding box'', $I_f$ is the image with bounding boxes, and $R_f$ is the corresponding caption. This approach allows the model to perceive fine-grained information, such as region-level details, within the image.

\begin{table*}[h!]
\centering
\caption{
Performance of different methods across different benchmarks. 
The best results in each column are made \textbf{bold}.
\textit{w-ECLIP}: LLaVA with enhanced CLIP trained on fine-grained data;
\textit{w-FineIns}: LLaVA trained on fine-grained visual instruction tuning data.
}
\resizebox{\linewidth}{!}{
\begin{tabular}{l|cc|cc|cc|cc|cc|cc|cc|cccc}
\toprule
\multirow{3}{*}{\textbf{Method}} & \multicolumn{6}{c|}{\textbf{POPE}} & \multicolumn{6}{c|}{\textbf{POPE-NoCaps}} & 
\multicolumn{4}{c}{\textbf{QA-VisualGenome}}\\ 
\cmidrule(lr){2-7} \cmidrule(lr){8-13} \cmidrule(lr){14-17} 
 & \multicolumn{2}{c|}{\textbf{Random}} & \multicolumn{2}{c|}{\textbf{Popular}} & \multicolumn{2}{c|}{\textbf{Adversarial}} & 
 \multicolumn{2}{c|}{\textbf{Random}} & \multicolumn{2}{c|}{\textbf{Popular}} & \multicolumn{2}{c|}{\textbf{Adversarial}} & \multicolumn{2}{c|}{\textbf{Attribute}}
 & \multicolumn{2}{c}{\textbf{Relation}}\\ 
\cmidrule(lr){2-17}
 & \textbf{Acc} & \textbf{F1} & \textbf{Acc} & \textbf{F1} & \textbf{Acc} & \textbf{F1} & \textbf{Acc} & \textbf{F1} & \textbf{Acc} & \textbf{F1} & \textbf{Acc} & \textbf{F1} & \textbf{Acc} & \textbf{F1}& \textbf{Acc} & \textbf{F1}\\ 
\midrule
\textit{LLaVA-7B} & 87.42 & 86.36 & 86.63 & 85.25 & 85.13 & 83.88 & 84.80 & 82.97 &79.40 &78.30 &74.77 &74.69 & 64.67 & 66.60 & 67.57 & 74.81\\ 

\textit{w-ECLIP} & \textbf{87.80} & \textbf{86.87} & \textbf{87.30} & \textbf{86.04} & \textbf{85.87} & \textbf{84.70} & 85.27 & 83.50 &81.00 &79.69 &75.77 &75.46 & 67.67 & 68.79 & 67.00 & 74.11\\
\textit{w-FineIns} & 87.77 & 86.78 & 86.80 &85.51 & 85.53 &84.33 & \textbf{85.53} & \textbf{84.00} & \textbf{81.73} & \textbf{80.61} & \textbf{76.50} & \textbf{76.37} & \textbf{69.01} & \textbf{70.12} & \textbf{69.75} & \textbf{76.17} \\
\bottomrule
\end{tabular}
\label{tab:clip_analysis}
}
\end{table*}

\subsection{How to reduce hallucination caused by the projector?}

In Section~\ref{sec:analysis_projector}, we reveal that hallucination introduced by 
the projector may be due to the inability of aligning visual and textual spaces, manifested by the low cosine similarity of caption embeddings and projected image features. 
Therefore, a straightforward remedy would be to explicitly bridge the image and caption representation during LLaVA's alignment stage.

\subsubsection{Loss Objectives}
Besides autoregressive image-text generation loss:
\begin{equation}
    \mathcal{L}_{itg} = -p( R | I, T ) 
\end{equation}
we introduce an in-batch contrastive alignment loss $\mathcal{L}_{itc}$ similar to Equation~\ref{eq:contrastive_loss}, where we maximize the similarity between a projected image feature and the corresponding text embedding for its caption.
\textit{We only focus on the alignment stage} and design three settings that involve the contrastive loss in different fashions.

\paragraph{Integrated Alignment Loss \includegraphics[height=1em]{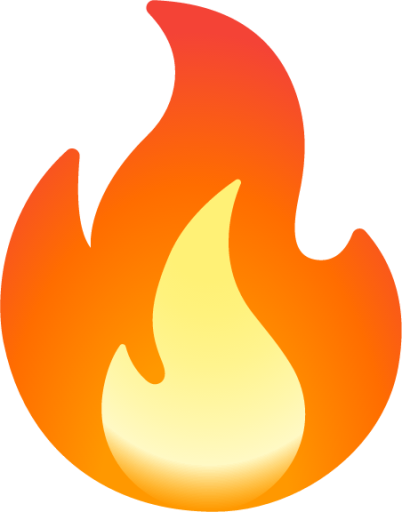}}\
The training process consists of two stages: \textit{alignment} and \textit{visual instruction tuning}.
The contrastive loss is integrated to the \textit{alignment} stage with a \textit{learnable} (\includegraphics[height=1em]{figures/fire_emoji.png}) weight $\lambda$.
The alignment objective is given by: 
$
\min_{{\mathcal{P}, \lambda}}   \mathcal{L}_{itg} + \lambda \mathcal{L}_{itc}
$. The visual instruction tuning stage is identical to LLaVA's.

\begin{table*}[ht]
\centering
\caption{
Performance of different projector alignment methods across different benchmarks.
The best results in each column are made \textbf{bold}. \textit{Int. Align.}: Integrated Alignment Loss with trainable (\includegraphics[height=1em]{figures/fire_emoji.png}) / frozen(\includegraphics[height=1em]{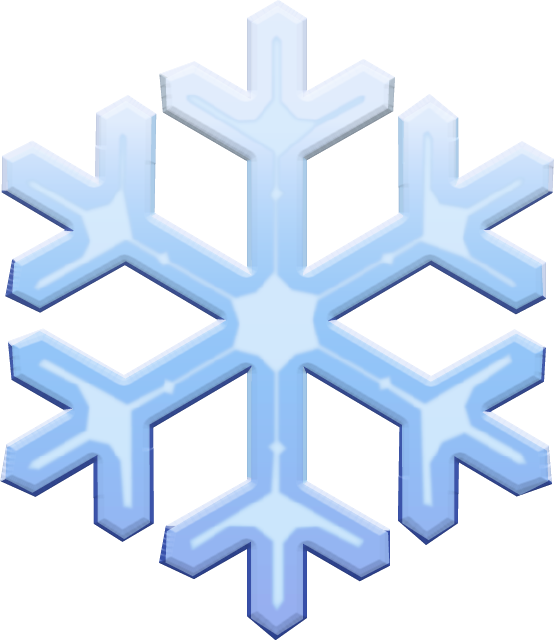}) weighting parameter; \textit{Sep. Ctrs. Align.}: Separate Contrastive Alignment Loss. }
\resizebox{\linewidth}{!}{
\begin{tabular}{l|cc|cc|cc|cc|cc|cc|cc|cc}
\toprule
\multirow{3}{*}{\textbf{Method}} & \multicolumn{6}{c|}{\textbf{POPE}} & \multicolumn{6}{c}{\textbf{POPE-NoCaps}}   & \multicolumn{4}{c}{\textbf{QA-VisualGenome}}\\ 
\cmidrule(lr){2-7} \cmidrule(lr){8-13} \cmidrule(lr){14-17}
 & \multicolumn{2}{c|}{\textbf{Random}} & \multicolumn{2}{c|}{\textbf{Popular}} & \multicolumn{2}{c|}{\textbf{Adversarial}} & 
 \multicolumn{2}{c|}{\textbf{Random}} & \multicolumn{2}{c|}{\textbf{Popular}} & \multicolumn{2}{c}{\textbf{Adversarial}} &  \multicolumn{2}{c|}{\textbf{Attribute}}
 & \multicolumn{2}{c}{\textbf{Relation}} \\ 
\cmidrule(lr){2-17}
 & \textbf{Acc} & \textbf{F1} & \textbf{Acc} & \textbf{F1} & \textbf{Acc} & \textbf{F1} & \textbf{Acc} & \textbf{F1} & \textbf{Acc} & \textbf{F1} & \textbf{Acc} & \textbf{F1} & \textbf{Acc} & \textbf{F1}& \textbf{Acc} & \textbf{F1}\\ 
\midrule
\textit{LLaVA-7B} & 87.42 & 86.36 & 86.63 & 85.25 & \textbf{85.13} & \textbf{83.88} & 84.80 & 82.97 & 79.40 & 78.30 & 74.77 & 74.69 & \textbf{64.67} & \textbf{66.60} & 67.57 & 74.81 \\
\textit{Int. Align.} \includegraphics[height=1em]{figures/fire_emoji.png} & 88.21 & 87.41 & 86.70 & 85.65 & 84.27 & 83.46 & \textbf{85.57} & \textbf{84.46} & 77.27 & 77.58 & 72.23 & 73.91 & 60.95 & 61.97 & 66.67 & 74.60\\
\textit{Int. Align.} \includegraphics[height=1em]{figures/snowflake_emoji.png} & 88.04 & 87.20 & 86.67 & 85.56 & 84.50 & 83.60 & 84.90 & 83.28 & 79.37 & 78.47 & 74.57 & 74.76 & 63.84 & 65.21 & 66.73 & 74.26\\
\textit{Sep. Ctrs. Align.} & \textbf{88.56} & \textbf{87.86} & \textbf{87.33} & \textbf{86.38} & 84.57 & \textbf{83.88} & \textbf{85.57} & 84.24 & \textbf{80.07} & \textbf{79.42} & \textbf{75.13} & \textbf{75.54} & 64.26 & 64.77 & \textbf{69.60} & \textbf{76.06}\\
\bottomrule
\end{tabular}
\label{tab:projector_analysis1}
}
\end{table*}

\paragraph{Integrated Alignment Loss \includegraphics[height=1em]{figures/snowflake_emoji.png}}
All settings are the same as above except that the weight $\lambda$ is \textit{fixed} (\includegraphics[height=1em]{figures/snowflake_emoji.png}).
The alignment objective is given by: 
$
\min_{{\mathcal{P}}}  \mathcal{L}_{itg} + \lambda \mathcal{L}_{itc}
$.

\paragraph{Separate Contrastive Alignment Loss}
We prepend a \textit{contrastive alignment stage} solely for the projector $\mathcal{P}$.
Namely, the first stage objective is given by: $\min_{{\mathcal{P}}}  \mathcal{L}_{itc}$.
The second stage and third stage correspond to the original \textit{autoregressive alignment} and \textit{visual instruction tuning} stage.

\section{Results and Analysis}

We first introduce the benchmarks on which our methods to be evaluated, which are shown as follows. 1) Object-based benchmarks: testing the object perception of LVLMs. \textit{POPE} and \textit{POPE-NoCaps}~\citep{DBLP:conf/emnlp/LiuJSWZ24} are adopted, where the latter is built on NoCaps~\cite{DBLP:conf/iccv/AgrawalAD0CJ0BP19} following a similar manner as in POPE.
2) Attribute- and relation-based benchmark: \textit{QA-VisualGenome} is constructed and adopted (detailed in Sec.~\ref{sec:analysis_settings}). 
We provide an in-depth analysis of our methods for improving the vision encoder and the projector. 
We call object-, attribute- and relation-based benchmarks as perception-based benchmarks.

For a fair comparison, we only use the LLaVA-Caption dataset for alignment.
All experiments are conducted on 4*A100 GPUs. For the \textit{alignment} stage, we set per-GPU batch size to 64, which is also the batch size contrastive alignment. 
We choose the well-known LLaVA-v1.5-7B model as our baseline.
All three settings introduce no extra learnable parameters (except for the weighting parameter $\lambda$ in \textbf{Integrated Alignment Loss} \includegraphics[height=1em]{figures/fire_emoji.png} setting).
Under our setting, both the \textit{original} and \textit{integrated alignment stage} take 6 hours, and \textit{visual instruction tuning stage} takes 24 hours.
Notably, the prepended \textit{contrastive alignment stage} takes only 12 minutes to train since only the vision encoder $\mathcal{V}$, projector $\mathcal{P}$ and the embedding layer of LLM $\mathcal{D}$ are involved in the forward process.
For the two integrated loss settings, we empirically initialize $\lambda$ with 5, make it learnable for \includegraphics[height=1em]{figures/fire_emoji.png} while keep it fixed for \includegraphics[height=1em]{figures/snowflake_emoji.png}. $\lambda_1$ and $\lambda_2$ are set to 1.

\subsection{Can our methods reduce hallucination caused by the vision encoder?}

Table \ref{tab:clip_analysis} presents the comprehensive experimental results of various settings across different testing benchmarks. From this table, several key observations can be drawn:
\textbf{1) Our proposed \textit{w-ECLIP} method demonstrates superior performance compared to LLaVA-7B on perception-based benchmarks}. This result underscores the effectiveness of our approach in reducing visual object hallucinations by enhancing the fine-grained perception capabilities of CLIP.
\textbf{2) \textit{w-FineIns} exhibits better performance than baseline on perception-based benchmarks}. This finding suggests that our fine-grained instruction data can augment the fine-grained perception abilities of LLaVA by leveraging region-level captions during training.
\textbf{3) Compared to \textit{w-FineIns}, \textit{w-ECLIP} demonstrates comparable or even better performance on perception-based benchmarks}. Notably, \textit{w-FineIns} offers efficiency advantages as it only requires the final training stage—instruction tuning—for the LVLM, simplifying the overall training process.

\subsection{Can our methods reduce hallucination caused by the projector?}

We benchmark our methods in Table~\ref{tab:projector_analysis1}.
For object-oriented benchmarks POPE and POPE-NoCaps, the model trained with \textit{Separate Contrastive Alignment Loss} outperforms others on most splits of benchmarks, though the improvement over baseline seems marginal. 
For QA-VisualGenome benchmark, we only observe improvement on the ``Relation'' split with \textit{Separate Contrastive Alignment Loss}, whereas slight performance drops are observed for others.
These observations provide insights for the alignment process.
Firstly, \textbf{object hallucinations may not be directly related to alignment in LVLM}, where vision encoder is mostly responsible for the perception process.
Secondly, \textbf{perception-based attribute and relation hallucination can hardly be mitigated by contrastive training of projector}. Similar to object hallucination, better visual representations may be needed as a remedy.


    

    

\subsection{Can our method influence other hallucinations?}

\begin{table}[]
\centering
\caption{Performance of different methods on QA-FB15K.}
\resizebox{0.9\linewidth}{!}{
\begin{tabular}{l|cc|cc}
\toprule
\multirow{2}{*}{\textbf{Method}} & \multicolumn{2}{c|}{\textbf{Entity}} & \multicolumn{2}{c}{\textbf{Relation}} \\ 
\cmidrule(lr){2-5}
 & \textbf{Acc} & \textbf{F1} & \textbf{Acc} & \textbf{F1} \\ 
\midrule
\textit{LLaVA-7B} & 78.39 & 73.14 & 56.79 & 48.79 \\ 
\textit{Int. Align.} \includegraphics[height=1em]{figures/fire_emoji.png} & \textbf{84.28} & \textbf{83.03} & 59.16 & \textbf{58.07} \\ 
\textit{Int. Align.} \includegraphics[height=1em]{figures/snowflake_emoji.png} & 84.05 & 81.76 & 59.16 & 56.97 \\ 
\textit{Sep. Ctrs. Align.} & 83.94 & 81.65 & \textbf{59.39} & 57.41 \\ 
\textit{LLaVA-7B} & 78.39 & 73.14 & 56.79 & 48.70 \\ 
\textit{w-ECLIP} & 77.60 & 71.47 & 56.79 & 45.58 \\ 
\textit{w-FineIns} & 76.47 & 69.86 & 55.45 & 49.10 \\ 
\bottomrule
\end{tabular}
}
\label{tab:qa_fb15k_align}
\end{table}

To further investigate the influence of our method on other kinds of hallucination, we introduced the Cognition-based benchmark: necessitating world knowledge in LVLMs for problem solving. We construct a cognition-based benchmark \textit{QA-FB15k} based on the knowledge graph FB15K \cite{DBLP:conf/nips/BordesUGWY13}. We show the results in Table \ref{tab:qa_fb15k_align}.

{Contrastive alignment objective is beneficial for cognition-based knowledge}, as evidenced by the performance boost on QA-FB15K. By better aligning between vision encoder and LLM, the LVLM is able to leverage the ability of LLM to answer the question that requires world knowledge, which is typically stored in LLMs pretrained on mountains of data. Nevertheless, performance boosts are found on QA-FB15K for all three settings over baselines.
{ Neither {w-FineIns} nor {w-ECLIP} shows any improvement on the cognition-based benchmark}. This may be attributed to the fact that, unlike perception-based benchmarks, cognition-based benchmarks necessitate not only the ability to identify objects but also the comprehension and application of relevant associated knowledge.  
The two methods primarily focus on improving perception, may not cater for the knowledge-intensive requirements of cognition-based benchmarks.

\textcolor{black}{\paragraph{More Analysis:} In addition, we add more experimental results on the hallucination benchmark and general benchmark, ablation study, and performance comparison with more baselines in Appendix \ref{app:hallucination_benchmark}, \ref{app:generalbenchmark},  \ref{app:ablation}, and \ref{app:comparison}.}


\section{Related Work}

Our work is related to the large vision-language model and hallucination in large vision-language model.

\paragraph{Large Vision-Language Model.} 
The multimodal learning field has recently pivoted its focus towards Large Vision-Language Models (LVLMs)~\cite{DBLP:journals/corr/abs-2308-01390,DBLP:journals/corr/abs-2305-03726}. Current advanced LVLMs primarily comprise three essential components: a language encoder, a visual encoder, and a cross-modal alignment mechanism~\cite{DBLP:conf/emnlp/RohrbachHBDS18}. Typically, the language encoder is implemented as a language model, such as LLaMA~\cite{llama} or Vicuna~\cite{vicuna}. In contrast, the visual encoder is usually based on VFMs like ViT~\cite{vit}. The role of the cross-modal alignment component is to integrate visual features with text representations, enabling the language encoder to effectively interpret visual semantics. 
To achieve comprehensive visual understanding, LVLMs generally undergo a series of training stages~\cite{MultiModal-GPT, minigpt4, llava15, llava, mPLUG-Owl, InstructBLIP}. For example, \citet{llava} align image features with word embeddings of a pre-trained language model during an initial alignment phase, followed by fine-tuning with tailored language-image instruction-following datasets. 
Despite these significant advancements, LVLMs continue to face challenges with hallucination, which significantly affects their performance across various vision-language applications.

\paragraph{Hallucinations in Large Vision-language Models.} 
Since hallucination issues and mitigation techniques have been extensively explored in text generation~\cite{ji2023survey, factscore}, research on hallucinations in LVLMs~\cite{InstructBLIP, llava, DBLP:journals/corr/abs-2404-05046} attracts more attention. 
To evaluate the hallucination in the LVLMs, several researchers propose metrics and benchmarks \cite{DBLP:conf/emnlp/RohrbachHBDS18,DBLP:conf/emnlp/LiDZWZW23,lovenia2023negative,DBLP:journals/corr/abs-2309-04041,jing2024faithscorefinegrainedevaluationshallucinations}.
Recently, various methods have been proposed to mitigate hallucinations in LVLMs, leveraging a range of techniques including decoding strategies~\citep{DBLP:journals/corr/abs-2311-16922, DBLP:journals/corr/abs-2311-17911}, post-processing methods~\citep{DBLP:journals/corr/abs-2310-00754,DBLP:journals/corr/abs-2409-16494,DBLP:journals/corr/abs-2310-16045}, the development of higher-quality datasets~\citep{DBLP:journals/corr/abs-2306-14565, DBLP:journals/corr/abs-2306-04387}, and modality alignment\cite{DBLP:journals/corr/abs-2312-10665, RLHF-V, DBLP:journals/corr/abs-2402-11411,DBLP:journals/corr/abs-2404-05046,2023llavarlhf,gunjal2023detecting}. 
Despite the success of the existing works, there lacks a comprehensive study of what causes visual hallucinations in LVLMs.



\section{Conclusion}
In this paper, our study delves into the visual hallucination problem in LVLMs, identifying its sources within the model's components. By independently analyzing the LLM, vision backbone, and projector, we propose targeted mitigation strategies. We introduce fine-grained hallucination benchmarks, QA-VisualGenome and QA-FB15k, to comprehensively evaluate hallucinations. Our methods demonstrate effectiveness in reducing hallucinations, contributing to the reliability and accuracy of LVLMs.

\section*{Limitations}
Our work primarily focuses on analyzing and improving hallucinations of general objects, such as tables and people, while neglecting the research topic of how to mitigate cognition-level hallucinations, such as the names of individuals and famous buildings.



\bibliography{custom}

\begin{thebibliography}{65}
\expandafter\ifx\csname natexlab\endcsname\relax\def\natexlab#1{#1}\fi

\bibitem[{Agrawal et~al.(2019)Agrawal, Anderson, Desai, Wang, Chen, Jain, Johnson, Batra, Parikh, and Lee}]{DBLP:conf/iccv/AgrawalAD0CJ0BP19}
Harsh Agrawal, Peter Anderson, Karan Desai, Yufei Wang, Xinlei Chen, Rishabh Jain, Mark Johnson, Dhruv Batra, Devi Parikh, and Stefan Lee. 2019.
\newblock \href {https://doi.org/10.1109/ICCV.2019.00904} {nocaps: novel object captioning at scale}.
\newblock In \emph{2019 {IEEE/CVF} International Conference on Computer Vision, {ICCV} 2019, Seoul, Korea (South), October 27 - November 2, 2019}, pages 8947--8956. {IEEE}.

\bibitem[{An et~al.(2025)An, Tian, Leng, Nie, Lin, Wang, Chen, Zhang, and Lu}]{an2025mitigatingobjecthallucinationslarge}
Wenbin An, Feng Tian, Sicong Leng, Jiahao Nie, Haonan Lin, QianYing Wang, Ping Chen, Xiaoqin Zhang, and Shijian Lu. 2025.
\newblock \href {http://arxiv.org/abs/2406.12718} {Mitigating object hallucinations in large vision-language models with assembly of global and local attention}.

\bibitem[{Antol et~al.(2015)Antol, Agrawal, Lu, Mitchell, Batra, Zitnick, and Parikh}]{vqatask}
Stanislaw Antol, Aishwarya Agrawal, Jiasen Lu, Margaret Mitchell, Dhruv Batra, C.~Lawrence Zitnick, and Devi Parikh. 2015.
\newblock {VQA:} visual question answering.
\newblock In \emph{{IEEE} International Conference on Computer Vision}, pages 2425--2433. {IEEE} Computer Society.

\bibitem[{Awadalla et~al.(2023)Awadalla, Gao, Gardner, Hessel, Hanafy, Zhu, Marathe, Bitton, Gadre, Sagawa, Jitsev, Kornblith, Koh, Ilharco, Wortsman, and Schmidt}]{DBLP:journals/corr/abs-2308-01390}
Anas Awadalla, Irena Gao, Josh Gardner, Jack Hessel, Yusuf Hanafy, Wanrong Zhu, Kalyani Marathe, Yonatan Bitton, Samir~Yitzhak Gadre, Shiori Sagawa, Jenia Jitsev, Simon Kornblith, Pang~Wei Koh, Gabriel Ilharco, Mitchell Wortsman, and Ludwig Schmidt. 2023.
\newblock Openflamingo: An open-source framework for training large autoregressive vision-language models.
\newblock \emph{CoRR}, abs/2308.01390.

\bibitem[{Bommasani et~al.(2021)Bommasani, Hudson, Adeli, Altman, Arora, von Arx, Bernstein, Bohg, Bosselut, Brunskill, Brynjolfsson, Buch, Card, Castellon, Chatterji, Chen, Creel, Davis, Demszky, Donahue, Doumbouya, Durmus, Ermon, Etchemendy, Ethayarajh, Fei{-}Fei, Finn, Gale, Gillespie, Goel, Goodman, Grossman, Guha, Hashimoto, Henderson, Hewitt, Ho, Hong, Hsu, Huang, Icard, Jain, Jurafsky, Kalluri, Karamcheti, Keeling, Khani, Khattab, Koh, Krass, Krishna, Kuditipudi, and et~al.}]{DBLP:journals/corr/abs-2108-07258}
Rishi Bommasani, Drew~A. Hudson, Ehsan Adeli, Russ~B. Altman, Simran Arora, Sydney von Arx, Michael~S. Bernstein, Jeannette Bohg, Antoine Bosselut, Emma Brunskill, Erik Brynjolfsson, Shyamal Buch, Dallas Card, Rodrigo Castellon, Niladri~S. Chatterji, Annie~S. Chen, Kathleen Creel, Jared~Quincy Davis, Dorottya Demszky, Chris Donahue, Moussa Doumbouya, Esin Durmus, Stefano Ermon, John Etchemendy, Kawin Ethayarajh, Li~Fei{-}Fei, Chelsea Finn, Trevor Gale, Lauren Gillespie, Karan Goel, Noah~D. Goodman, Shelby Grossman, Neel Guha, Tatsunori Hashimoto, Peter Henderson, John Hewitt, Daniel~E. Ho, Jenny Hong, Kyle Hsu, Jing Huang, Thomas Icard, Saahil Jain, Dan Jurafsky, Pratyusha Kalluri, Siddharth Karamcheti, Geoff Keeling, Fereshte Khani, Omar Khattab, Pang~Wei Koh, Mark~S. Krass, Ranjay Krishna, Rohith Kuditipudi, and et~al. 2021.
\newblock \href {http://arxiv.org/abs/2108.07258} {On the opportunities and risks of foundation models}.
\newblock \emph{CoRR}, abs/2108.07258.

\bibitem[{Bordes et~al.(2013)Bordes, Usunier, Garc{\'{\i}}a{-}Dur{\'{a}}n, Weston, and Yakhnenko}]{DBLP:conf/nips/BordesUGWY13}
Antoine Bordes, Nicolas Usunier, Alberto Garc{\'{\i}}a{-}Dur{\'{a}}n, Jason Weston, and Oksana Yakhnenko. 2013.
\newblock \href {https://proceedings.neurips.cc/paper/2013/hash/1cecc7a77928ca8133fa24680a88d2f9-Abstract.html} {Translating embeddings for modeling multi-relational data}.
\newblock In \emph{Advances in Neural Information Processing Systems 26: 27th Annual Conference on Neural Information Processing Systems 2013. Proceedings of a meeting held December 5-8, 2013, Lake Tahoe, Nevada, United States}, pages 2787--2795.

\bibitem[{Brown(2020)}]{gpt3}
Tom~B Brown. 2020.
\newblock Language models are few-shot learners.
\newblock \emph{arXiv preprint arXiv:2005.14165}.

\bibitem[{Chang et~al.(2024)Chang, Jing, Zhang, and Zhang}]{DBLP:journals/corr/abs-2409-16494}
Yue Chang, Liqiang Jing, Xiaopeng Zhang, and Yue Zhang. 2024.
\newblock \href {https://doi.org/10.48550/ARXIV.2409.16494} {A unified hallucination mitigation framework for large vision-language models}.
\newblock \emph{CoRR}, abs/2409.16494.

\bibitem[{Chen et~al.(2024{\natexlab{a}})Chen, Chen, Zhang, Chen, Wu, Zhang, Chen, Li, Wan, and Wang}]{chen2024allava}
Guiming~Hardy Chen, Shunian Chen, Ruifei Zhang, Junying Chen, Xiangbo Wu, Zhiyi Zhang, Zhihong Chen, Jianquan Li, Xiang Wan, and Benyou Wang. 2024{\natexlab{a}}.
\newblock Allava: Harnessing gpt4v-synthesized data for a lite vision-language model.
\newblock \emph{arXiv preprint arXiv:2402.11684}.

\bibitem[{Chen et~al.(2024{\natexlab{b}})Chen, Zhao, Liu, Bai, Lin, Zhou, and Chang}]{fastv}
Liang Chen, Haozhe Zhao, Tianyu Liu, Shuai Bai, Junyang Lin, Chang Zhou, and Baobao Chang. 2024{\natexlab{b}}.
\newblock \href {https://doi.org/10.1007/978-3-031-73004-7\_2} {An image is worth 1/2 tokens after layer 2: Plug-and-play inference acceleration for large vision-language models}.
\newblock In \emph{Computer Vision - {ECCV} 2024 - 18th European Conference, Milan, Italy, September 29-October 4, 2024, Proceedings, Part {LXXXI}}, volume 15139 of \emph{Lecture Notes in Computer Science}, pages 19--35. Springer.

\bibitem[{Chen et~al.(2024{\natexlab{c}})Chen, Zhao, Luo, Yao, Li, and Zhou}]{halc}
Zhaorun Chen, Zhuokai Zhao, Hongyin Luo, Huaxiu Yao, Bo~Li, and Jiawei Zhou. 2024{\natexlab{c}}.
\newblock \href {http://arxiv.org/abs/2403.00425} {Halc: Object hallucination reduction via adaptive focal-contrast decoding}.

\bibitem[{Chiang et~al.(2023)Chiang, Li, Lin, Sheng, Wu, Zhang, Zheng, Zhuang, Zhuang, Gonzalez, Stoica, and Xing}]{vicuna}
Wei-Lin Chiang, Zhuohan Li, Zi~Lin, Ying Sheng, Zhanghao Wu, Hao Zhang, Lianmin Zheng, Siyuan Zhuang, Yonghao Zhuang, Joseph~E. Gonzalez, Ion Stoica, and Eric~P. Xing. 2023.
\newblock vicuna: An opensource chatbot impressing gpt-4 with 90

\bibitem[{Chuang et~al.(2024)Chuang, Xie, Luo, Kim, Glass, and He}]{DBLP:conf/iclr/ChuangXLKGH24}
Yung{-}Sung Chuang, Yujia Xie, Hongyin Luo, Yoon Kim, James~R. Glass, and Pengcheng He. 2024.
\newblock \href {https://openreview.net/forum?id=Th6NyL07na} {Dola: Decoding by contrasting layers improves factuality in large language models}.
\newblock In \emph{The Twelfth International Conference on Learning Representations, {ICLR} 2024, Vienna, Austria, May 7-11, 2024}. OpenReview.net.

\bibitem[{Dai et~al.(2023)Dai, Li, Li, Tiong, Zhao, Wang, Li, Fung, and Hoi}]{InstructBLIP}
Wenliang Dai, Junnan Li, Dongxu Li, Anthony Meng~Huat Tiong, Junqi Zhao, Weisheng Wang, Boyang Li, Pascale Fung, and Steven C.~H. Hoi. 2023.
\newblock Instructblip: Towards general-purpose vision-language models with instruction tuning.
\newblock \emph{CoRR}, abs/2305.06500.

\bibitem[{Deng et~al.(2009)Deng, Dong, Socher, Li, Li, and Fei-Fei}]{deng2009imagenet}
Jia Deng, Wei Dong, Richard Socher, Li-Jia Li, Kai Li, and Li~Fei-Fei. 2009.
\newblock Imagenet: A large-scale hierarchical image database.
\newblock In \emph{2009 IEEE conference on computer vision and pattern recognition}, pages 248--255. Ieee.

\bibitem[{Dosovitskiy et~al.(2021)Dosovitskiy, Beyer, Kolesnikov, Weissenborn, Zhai, Unterthiner, Dehghani, Minderer, Heigold, Gelly, Uszkoreit, and Houlsby}]{vit}
Alexey Dosovitskiy, Lucas Beyer, Alexander Kolesnikov, Dirk Weissenborn, Xiaohua Zhai, Thomas Unterthiner, Mostafa Dehghani, Matthias Minderer, Georg Heigold, Sylvain Gelly, Jakob Uszkoreit, and Neil Houlsby. 2021.
\newblock An image is worth 16x16 words: Transformers for image recognition at scale.
\newblock In \emph{{ICLR}}. OpenReview.net.

\bibitem[{Gong et~al.(2023)Gong, Lyu, Zhang, Wang, Zheng, Zhao, Liu, Zhang, Luo, and Chen}]{MultiModal-GPT}
Tao Gong, Chengqi Lyu, Shilong Zhang, Yudong Wang, Miao Zheng, Qian Zhao, Kuikun Liu, Wenwei Zhang, Ping Luo, and Kai Chen. 2023.
\newblock Multimodal-gpt: {A} vision and language model for dialogue with humans.
\newblock \emph{CoRR}, abs/2305.04790.

\bibitem[{Gunjal et~al.(2023)Gunjal, Yin, and Bas}]{gunjal2023detecting}
Anisha Gunjal, Jihan Yin, and Erhan Bas. 2023.
\newblock Detecting and preventing hallucinations in large vision language models.
\newblock \emph{arXiv preprint arXiv:2308.06394}.

\bibitem[{Hewitt et~al.(2021)Hewitt, Ethayarajh, Liang, and Manning}]{hewitt2021conditional}
John Hewitt, Kawin Ethayarajh, Percy Liang, and Christopher~D Manning. 2021.
\newblock Conditional probing: measuring usable information beyond a baseline.
\newblock \emph{arXiv preprint arXiv:2109.09234}.

\bibitem[{Huang et~al.(2023)Huang, Dong, Zhang, Wang, He, Wang, Lin, Zhang, and Yu}]{DBLP:journals/corr/abs-2311-17911}
Qidong Huang, Xiaoyi Dong, Pan Zhang, Bin Wang, Conghui He, Jiaqi Wang, Dahua Lin, Weiming Zhang, and Nenghai Yu. 2023.
\newblock \href {https://doi.org/10.48550/ARXIV.2311.17911} {{OPERA:} alleviating hallucination in multi-modal large language models via over-trust penalty and retrospection-allocation}.
\newblock \emph{CoRR}, abs/2311.17911.

\bibitem[{Huang et~al.(2024{\natexlab{a}})Huang, Dong, Zhang, Wang, He, Wang, Lin, Zhang, and Yu}]{huang2024operaalleviatinghallucinationmultimodal}
Qidong Huang, Xiaoyi Dong, Pan Zhang, Bin Wang, Conghui He, Jiaqi Wang, Dahua Lin, Weiming Zhang, and Nenghai Yu. 2024{\natexlab{a}}.
\newblock \href {http://arxiv.org/abs/2311.17911} {Opera: Alleviating hallucination in multi-modal large language models via over-trust penalty and retrospection-allocation}.

\bibitem[{Huang et~al.(2024{\natexlab{b}})Huang, Dong, Zhang, Zang, Cao, Wang, Lin, Zhang, and Yu}]{huang2024deciphering}
Qidong Huang, Xiaoyi Dong, Pan Zhang, Yuhang Zang, Yuhang Cao, Jiaqi Wang, Dahua Lin, Weiming Zhang, and Nenghai Yu. 2024{\natexlab{b}}.
\newblock Deciphering cross-modal alignment in large vision-language models with modality integration rate.
\newblock \emph{arXiv preprint arXiv:2410.07167}.

\bibitem[{Ji et~al.(2023)Ji, Lee, Frieske, Yu, Su, Xu, Ishii, Bang, Madotto, and Fung}]{ji2023survey}
Ziwei Ji, Nayeon Lee, Rita Frieske, Tiezheng Yu, Dan Su, Yan Xu, Etsuko Ishii, Ye~Jin Bang, Andrea Madotto, and Pascale Fung. 2023.
\newblock Survey of hallucination in natural language generation.
\newblock \emph{ACM Computing Surveys}, 55(12):1--38.

\bibitem[{Jing and Du(2024)}]{DBLP:journals/corr/abs-2404-05046}
Liqiang Jing and Xinya Du. 2024.
\newblock \href {https://doi.org/10.48550/ARXIV.2404.05046} {{FGAIF:} aligning large vision-language models with fine-grained {AI} feedback}.
\newblock \emph{CoRR}, abs/2404.05046.

\bibitem[{Jing et~al.(2024)Jing, Li, Chen, and Du}]{jing2024faithscorefinegrainedevaluationshallucinations}
Liqiang Jing, Ruosen Li, Yunmo Chen, and Xinya Du. 2024.
\newblock \href {http://arxiv.org/abs/2311.01477} {Faithscore: Fine-grained evaluations of hallucinations in large vision-language models}.

\bibitem[{Krishna et~al.(2017)Krishna, Zhu, Groth, Johnson, Hata, Kravitz, Chen, Kalantidis, Li, Shamma, Bernstein, and Fei{-}Fei}]{DBLP:journals/ijcv/KrishnaZGJHKCKL17}
Ranjay Krishna, Yuke Zhu, Oliver Groth, Justin Johnson, Kenji Hata, Joshua Kravitz, Stephanie Chen, Yannis Kalantidis, Li{-}Jia Li, David~A. Shamma, Michael~S. Bernstein, and Li~Fei{-}Fei. 2017.
\newblock \href {https://doi.org/10.1007/S11263-016-0981-7} {Visual genome: Connecting language and vision using crowdsourced dense image annotations}.
\newblock \emph{Int. J. Comput. Vis.}, 123(1):32--73.

\bibitem[{Krizhevsky et~al.(2009)Krizhevsky, Hinton et~al.}]{krizhevsky2009learning}
Alex Krizhevsky, Geoffrey Hinton, et~al. 2009.
\newblock Learning multiple layers of features from tiny images.

\bibitem[{Krizhevsky et~al.()Krizhevsky, Nair, and Hinton}]{cifar100}
Alex Krizhevsky, Vinod Nair, and Geoffrey Hinton.
\newblock \href {http://www.cs.toronto.edu/~kriz/cifar.html} {Cifar-100 (canadian institute for advanced research)}.

\bibitem[{Leng et~al.(2023)Leng, Zhang, Chen, Li, Lu, Miao, and Bing}]{DBLP:journals/corr/abs-2311-16922}
Sicong Leng, Hang Zhang, Guanzheng Chen, Xin Li, Shijian Lu, Chunyan Miao, and Lidong Bing. 2023.
\newblock \href {https://doi.org/10.48550/ARXIV.2311.16922} {Mitigating object hallucinations in large vision-language models through visual contrastive decoding}.
\newblock \emph{CoRR}, abs/2311.16922.

\bibitem[{Li et~al.(2023{\natexlab{a}})Li, Zhang, Chen, Wang, Yang, and Liu}]{DBLP:journals/corr/abs-2305-03726}
Bo~Li, Yuanhan Zhang, Liangyu Chen, Jinghao Wang, Jingkang Yang, and Ziwei Liu. 2023{\natexlab{a}}.
\newblock Otter: {A} multi-modal model with in-context instruction tuning.
\newblock \emph{CoRR}, abs/2305.03726.

\bibitem[{Li et~al.(2023{\natexlab{b}})Li, Patel, Vi{\'{e}}gas, Pfister, and Wattenberg}]{DBLP:conf/nips/0002PVPW23}
Kenneth Li, Oam Patel, Fernanda~B. Vi{\'{e}}gas, Hanspeter Pfister, and Martin Wattenberg. 2023{\natexlab{b}}.
\newblock \href {http://papers.nips.cc/paper\_files/paper/2023/hash/81b8390039b7302c909cb769f8b6cd93-Abstract-Conference.html} {Inference-time intervention: Eliciting truthful answers from a language model}.
\newblock In \emph{Advances in Neural Information Processing Systems 36: Annual Conference on Neural Information Processing Systems 2023, NeurIPS 2023, New Orleans, LA, USA, December 10 - 16, 2023}.

\bibitem[{Li et~al.(2023{\natexlab{c}})Li, Xie, Li, Chen, Wang, Chen, Yang, Wang, and Kong}]{DBLP:journals/corr/abs-2312-10665}
Lei Li, Zhihui Xie, Mukai Li, Shunian Chen, Peiyi Wang, Liang Chen, Yazheng Yang, Benyou Wang, and Lingpeng Kong. 2023{\natexlab{c}}.
\newblock \href {https://doi.org/10.48550/ARXIV.2312.10665} {Silkie: Preference distillation for large visual language models}.
\newblock \emph{CoRR}, abs/2312.10665.

\bibitem[{Li et~al.(2023{\natexlab{d}})Li, Yin, Li, Chen, Wang, Ren, Li, Yang, Xu, Sun, Kong, and Liu}]{DBLP:journals/corr/abs-2306-04387}
Lei Li, Yuwei Yin, Shicheng Li, Liang Chen, Peiyi Wang, Shuhuai Ren, Mukai Li, Yazheng Yang, Jingjing Xu, Xu~Sun, Lingpeng Kong, and Qi~Liu. 2023{\natexlab{d}}.
\newblock \href {https://doi.org/10.48550/ARXIV.2306.04387} {M\({}^{\mbox{3}}\)it: {A} large-scale dataset towards multi-modal multilingual instruction tuning}.
\newblock \emph{CoRR}, abs/2306.04387.

\bibitem[{Li et~al.(2025)Li, Geng, Zhu, Chen, Song, Ma, and Karray}]{DBLP:conf/aaai/LiGZCSMK25}
Qing Li, Jiahui Geng, Derui Zhu, Zongxiong Chen, Kun Song, Lei Ma, and Fakhri Karray. 2025.
\newblock \href {https://doi.org/10.1609/AAAI.V39I26.34954} {Internal activation revision: Safeguarding vision language models without parameter update}.
\newblock In \emph{AAAI-25, Sponsored by the Association for the Advancement of Artificial Intelligence, February 25 - March 4, 2025, Philadelphia, PA, {USA}}, pages 27428--27436. {AAAI} Press.

\bibitem[{Li et~al.(2023{\natexlab{e}})Li, Du, Zhou, Wang, Zhao, and Wen}]{DBLP:conf/emnlp/LiDZWZW23}
Yifan Li, Yifan Du, Kun Zhou, Jinpeng Wang, Wayne~Xin Zhao, and Ji{-}Rong Wen. 2023{\natexlab{e}}.
\newblock \href {https://doi.org/10.18653/V1/2023.EMNLP-MAIN.20} {Evaluating object hallucination in large vision-language models}.
\newblock In \emph{Proceedings of the 2023 Conference on Empirical Methods in Natural Language Processing, {EMNLP} 2023, Singapore, December 6-10, 2023}, pages 292--305. Association for Computational Linguistics.

\bibitem[{Lin et~al.(2014)Lin, Maire, Belongie, Hays, Perona, Ramanan, Doll{\'{a}}r, and Zitnick}]{mscoco}
Tsung{-}Yi Lin, Michael Maire, Serge~J. Belongie, James Hays, Pietro Perona, Deva Ramanan, Piotr Doll{\'{a}}r, and C.~Lawrence Zitnick. 2014.
\newblock Microsoft {COCO:} common objects in context.
\newblock In \emph{ECCV}, volume 8693 of \emph{Lecture Notes in Computer Science}, pages 740--755. Springer.

\bibitem[{Liu et~al.(2023{\natexlab{a}})Liu, Guan, Li, Chen, Yacoob, Manocha, and Zhou}]{liu2023hallusionbench}
Fuxiao Liu, Tianrui Guan, Zongxia Li, Lichang Chen, Yaser Yacoob, Dinesh Manocha, and Tianyi Zhou. 2023{\natexlab{a}}.
\newblock Hallusionbench: You see what you think? or you think what you see? an image-context reasoning benchmark challenging for gpt-4v (ision), llava-1.5, and other multi-modality models.
\newblock \emph{arXiv preprint arXiv:2310.14566}.

\bibitem[{Liu et~al.(2023{\natexlab{b}})Liu, Lin, Li, Wang, Yacoob, and Wang}]{liu2023aligning}
Fuxiao Liu, Kevin Lin, Linjie Li, Jianfeng Wang, Yaser Yacoob, and Lijuan Wang. 2023{\natexlab{b}}.
\newblock Aligning large multi-modal model with robust instruction tuning.
\newblock \emph{arXiv preprint arXiv:2306.14565}.

\bibitem[{Liu et~al.(2023{\natexlab{c}})Liu, Lin, Li, Wang, Yacoob, and Wang}]{DBLP:journals/corr/abs-2306-14565}
Fuxiao Liu, Kevin Lin, Linjie Li, Jianfeng Wang, Yaser Yacoob, and Lijuan Wang. 2023{\natexlab{c}}.
\newblock Aligning large multi-modal model with robust instruction tuning.
\newblock \emph{CoRR}, abs/2306.14565.

\bibitem[{Liu et~al.(2024{\natexlab{a}})Liu, Lin, Li, Wang, Yacoob, and Wang}]{liu2024mitigating}
Fuxiao Liu, Kevin Lin, Linjie Li, Jianfeng Wang, Yaser Yacoob, and Lijuan Wang. 2024{\natexlab{a}}.
\newblock \href {http://arxiv.org/abs/2306.14565} {Mitigating hallucination in large multi-modal models via robust instruction tuning}.

\bibitem[{Liu et~al.(2023{\natexlab{d}})Liu, Li, Li, and Lee}]{llava15}
Haotian Liu, Chunyuan Li, Yuheng Li, and Yong~Jae Lee. 2023{\natexlab{d}}.
\newblock Improved baselines with visual instruction tuning.
\newblock \emph{arXiv preprint arXiv:2310.03744}.

\bibitem[{Liu et~al.(2023{\natexlab{e}})Liu, Li, Wu, and Lee}]{llava}
Haotian Liu, Chunyuan Li, Qingyang Wu, and Yong~Jae Lee. 2023{\natexlab{e}}.
\newblock Visual instruction tuning.
\newblock \emph{CoRR}, abs/2304.08485.

\bibitem[{Liu et~al.(2024{\natexlab{b}})Liu, Ji, Sun, Wu, and Zhou}]{DBLP:conf/emnlp/LiuJSWZ24}
Yufang Liu, Tao Ji, Changzhi Sun, Yuanbin Wu, and Aimin Zhou. 2024{\natexlab{b}}.
\newblock \href {https://aclanthology.org/2024.emnlp-main.1016} {Investigating and mitigating object hallucinations in pretrained vision-language {(CLIP)} models}.
\newblock In \emph{Proceedings of the 2024 Conference on Empirical Methods in Natural Language Processing, {EMNLP} 2024, Miami, FL, USA, November 12-16, 2024}, pages 18288--18301. Association for Computational Linguistics.

\bibitem[{Lovenia et~al.(2023)Lovenia, Dai, Cahyawijaya, Ji, and Fung}]{lovenia2023negative}
Holy Lovenia, Wenliang Dai, Samuel Cahyawijaya, Ziwei Ji, and Pascale Fung. 2023.
\newblock \href {http://arxiv.org/abs/2310.05338} {Negative object presence evaluation (nope) to measure object hallucination in vision-language models}.

\bibitem[{Lu et~al.(2023)Lu, Rao, Chen, Guo, Zhang, Sun, Yang, and Yang}]{DBLP:journals/corr/abs-2309-04041}
Jiaying Lu, Jinmeng Rao, Kezhen Chen, Xiaoyuan Guo, Yawen Zhang, Baochen Sun, Carl Yang, and Jie Yang. 2023.
\newblock Evaluation and mitigation of agnosia in multimodal large language models.
\newblock \emph{CoRR}, abs/2309.04041.

\bibitem[{Min et~al.(2023)Min, Krishna, Lyu, Lewis, Yih, Koh, Iyyer, Zettlemoyer, and Hajishirzi}]{factscore}
Sewon Min, Kalpesh Krishna, Xinxi Lyu, Mike Lewis, Wen{-}tau Yih, Pang~Wei Koh, Mohit Iyyer, Luke Zettlemoyer, and Hannaneh Hajishirzi. 2023.
\newblock Factscore: Fine-grained atomic evaluation of factual precision in long form text generation.
\newblock \emph{CoRR}, abs/2305.14251.

\bibitem[{OpenAI(2022)}]{chatgpt}
OpenAI. 2022.
\newblock Chatgpt blog post.

\bibitem[{Rohrbach et~al.(2018)Rohrbach, Hendricks, Burns, Darrell, and Saenko}]{DBLP:conf/emnlp/RohrbachHBDS18}
Anna Rohrbach, Lisa~Anne Hendricks, Kaylee Burns, Trevor Darrell, and Kate Saenko. 2018.
\newblock Object hallucination in image captioning.
\newblock In \emph{EMNLP}, pages 4035--4045. ACL.

\bibitem[{Sun et~al.(2023{\natexlab{a}})Sun, Shen, Cao, Liu, Li, Shen, Gan, Gui, Wang, Yang, Keutzer, and Darrell}]{2023llavarlhf}
Zhiqing Sun, Sheng Shen, Shengcao Cao, Haotian Liu, Chunyuan Li, Yikang Shen, Chuang Gan, Liang{-}Yan Gui, Yu{-}Xiong Wang, Yiming Yang, Kurt Keutzer, and Trevor Darrell. 2023{\natexlab{a}}.
\newblock \href {https://doi.org/10.48550/ARXIV.2309.14525} {Aligning large multimodal models with factually augmented {RLHF}}.
\newblock \emph{CoRR}, abs/2309.14525.

\bibitem[{Sun et~al.(2023{\natexlab{b}})Sun, Shen, Cao, Liu, Shen, Gan, Gui, Wang, Yang, Keutzer, and Darrell}]{llava_rlhf}
Zhiqing Sun, Sheng Shen, Shengcao Cao, Haotian Liu, Yikang Shen, Chuang Gan, Liang{-}Yan Gui, Yu{-}Xiong Wang, Yiming Yang, Kurt Keutzer, and Trevor Darrell. 2023{\natexlab{b}}.
\newblock Aligning large multimodal models with factually augmented {RLHF}.
\newblock \emph{CoRR}, abs/2309.14525.

\bibitem[{Touvron et~al.(2023)Touvron, Lavril, Izacard, Martinet, Lachaux, Lacroix, Rozi{\`{e}}re, Goyal, Hambro, Azhar, Rodriguez, Joulin, Grave, and Lample}]{llama}
Hugo Touvron, Thibaut Lavril, Gautier Izacard, Xavier Martinet, Marie{-}Anne Lachaux, Timoth{\'{e}}e Lacroix, Baptiste Rozi{\`{e}}re, Naman Goyal, Eric Hambro, Faisal Azhar, Aur{\'{e}}lien Rodriguez, Armand Joulin, Edouard Grave, and Guillaume Lample. 2023.
\newblock Llama: Open and efficient foundation language models.
\newblock \emph{CoRR}, abs/2302.13971.

\bibitem[{Wang et~al.(2023)Wang, Wang, Xu, Zhang, Gu, Jia, Yan, Zhang, and Sang}]{DBLP:journals/corr/abs-2311-07397}
Junyang Wang, Yuhang Wang, Guohai Xu, Jing Zhang, Yukai Gu, Haitao Jia, Ming Yan, Ji~Zhang, and Jitao Sang. 2023.
\newblock \href {https://doi.org/10.48550/ARXIV.2311.07397} {An llm-free multi-dimensional benchmark for mllms hallucination evaluation}.
\newblock \emph{CoRR}, abs/2311.07397.

\bibitem[{Wang et~al.(2020)Wang, Huang, Zhang, and Sun}]{DBLP:conf/cvpr/WangHZS20a}
Tan Wang, Jianqiang Huang, Hanwang Zhang, and Qianru Sun. 2020.
\newblock \href {https://doi.org/10.1109/CVPR42600.2020.01077} {Visual commonsense {R-CNN}}.
\newblock In \emph{2020 {IEEE/CVF} Conference on Computer Vision and Pattern Recognition, {CVPR} 2020, Seattle, WA, USA, June 13-19, 2020}, pages 10757--10767. Computer Vision Foundation / {IEEE}.

\bibitem[{Wei and Zhang(2024)}]{DBLP:conf/mm/WeiZ24}
Jinfeng Wei and Xiaofeng Zhang. 2024.
\newblock \href {https://doi.org/10.1145/3664647.3681076} {{DOPRA:} decoding over-accumulation penalization and re-allocation in specific weighting layer}.
\newblock In \emph{Proceedings of the 32nd {ACM} International Conference on Multimedia, {MM} 2024, Melbourne, VIC, Australia, 28 October 2024 - 1 November 2024}, pages 7065--7074. {ACM}.

\bibitem[{Xing et~al.(2024)Xing, Li, Laptev, and Lu}]{DBLP:conf/nips/XingLLL24}
Yun Xing, Yiheng Li, Ivan Laptev, and Shijian Lu. 2024.
\newblock \href {http://papers.nips.cc/paper\_files/paper/2024/hash/a76ed4a8ef522c823d73925e7fff16d4-Abstract-Conference.html} {Mitigating object hallucination via concentric causal attention}.
\newblock In \emph{Advances in Neural Information Processing Systems 38: Annual Conference on Neural Information Processing Systems 2024, NeurIPS 2024, Vancouver, BC, Canada, December 10 - 15, 2024}.

\bibitem[{Ye et~al.(2023)Ye, Xu, Xu, Ye, Yan, Zhou, Wang, Hu, Shi, Shi, Li, Xu, Chen, Tian, Qi, Zhang, and Huang}]{mPLUG-Owl}
Qinghao Ye, Haiyang Xu, Guohai Xu, Jiabo Ye, Ming Yan, Yiyang Zhou, Junyang Wang, Anwen Hu, Pengcheng Shi, Yaya Shi, Chenliang Li, Yuanhong Xu, Hehong Chen, Junfeng Tian, Qian Qi, Ji~Zhang, and Fei Huang. 2023.
\newblock mplug-owl: Modularization empowers large language models with multimodality.
\newblock \emph{CoRR}, abs/2304.14178.

\bibitem[{Yin et~al.(2023)Yin, Fu, Zhao, Xu, Wang, Sui, Shen, Li, Sun, and Chen}]{DBLP:journals/corr/abs-2310-16045}
Shukang Yin, Chaoyou Fu, Sirui Zhao, Tong Xu, Hao Wang, Dianbo Sui, Yunhang Shen, Ke~Li, Xing Sun, and Enhong Chen. 2023.
\newblock Woodpecker: Hallucination correction for multimodal large language models.
\newblock \emph{CoRR}, abs/2310.16045.

\bibitem[{Yu et~al.(2023)Yu, Yao, Zhang, He, Han, Cui, Hu, Liu, Zheng, Sun, and Chua}]{RLHF-V}
Tianyu Yu, Yuan Yao, Haoye Zhang, Taiwen He, Yifeng Han, Ganqu Cui, Jinyi Hu, Zhiyuan Liu, Hai{-}Tao Zheng, Maosong Sun, and Tat{-}Seng Chua. 2023.
\newblock \href {https://doi.org/10.48550/ARXIV.2312.00849} {{RLHF-V:} towards trustworthy mllms via behavior alignment from fine-grained correctional human feedback}.
\newblock \emph{CoRR}, abs/2312.00849.

\bibitem[{Yue et~al.(2024)Yue, Zhang, and Jin}]{lessismore}
Zihao Yue, Liang Zhang, and Qin Jin. 2024.
\newblock \href {https://doi.org/10.18653/V1/2024.ACL-LONG.633} {Less is more: Mitigating multimodal hallucination from an {EOS} decision perspective}.
\newblock In \emph{Proceedings of the 62nd Annual Meeting of the Association for Computational Linguistics (Volume 1: Long Papers), {ACL} 2024, Bangkok, Thailand, August 11-16, 2024}, pages 11766--11781. Association for Computational Linguistics.

\bibitem[{Zhang et~al.(2024{\natexlab{a}})Zhang, Quan, Gu, Shen, Yuan, Yan, Cheng, Wu, and Ye}]{eah}
Xiaofeng Zhang, Yihao Quan, Chaochen Gu, Chen Shen, Xiaosong Yuan, Shaotian Yan, Hao Cheng, Kaijie Wu, and Jieping Ye. 2024{\natexlab{a}}.
\newblock \href {https://doi.org/10.48550/ARXIV.2411.09968} {Seeing clearly by layer two: Enhancing attention heads to alleviate hallucination in lvlms}.
\newblock \emph{CoRR}, abs/2411.09968.

\bibitem[{Zhang et~al.(2025)Zhang, Jing, and Gogate}]{DBLP:conf/aaai/ZhangJG25}
Yue Zhang, Liqiang Jing, and Vibhav Gogate. 2025.
\newblock \href {https://doi.org/10.1609/AAAI.V39I24.34792} {Defeasible visual entailment: Benchmark, evaluator, and reward-driven optimization}.
\newblock In \emph{AAAI-25, Sponsored by the Association for the Advancement of Artificial Intelligence, February 25 - March 4, 2025, Philadelphia, PA, {USA}}, pages 25976--25984. {AAAI} Press.

\bibitem[{Zhang et~al.(2024{\natexlab{b}})Zhang, Zuo, and Jing}]{zhang2024fine}
Yue Zhang, Jingxuan Zuo, and Liqiang Jing. 2024{\natexlab{b}}.
\newblock Fine-grained and explainable factuality evaluation for multimodal summarization.
\newblock \emph{arXiv preprint arXiv:2402.11414}.

\bibitem[{Zhou et~al.(2024)Zhou, Cui, Rafailov, Finn, and Yao}]{DBLP:journals/corr/abs-2402-11411}
Yiyang Zhou, Chenhang Cui, Rafael Rafailov, Chelsea Finn, and Huaxiu Yao. 2024.
\newblock \href {https://doi.org/10.48550/ARXIV.2402.11411} {Aligning modalities in vision large language models via preference fine-tuning}.
\newblock \emph{CoRR}, abs/2402.11411.

\bibitem[{Zhou et~al.(2023)Zhou, Cui, Yoon, Zhang, Deng, Finn, Bansal, and Yao}]{DBLP:journals/corr/abs-2310-00754}
Yiyang Zhou, Chenhang Cui, Jaehong Yoon, Linjun Zhang, Zhun Deng, Chelsea Finn, Mohit Bansal, and Huaxiu Yao. 2023.
\newblock \href {https://doi.org/10.48550/ARXIV.2310.00754} {Analyzing and mitigating object hallucination in large vision-language models}.
\newblock \emph{CoRR}, abs/2310.00754.

\bibitem[{Zhu et~al.(2023)Zhu, Chen, Shen, Li, and Elhoseiny}]{minigpt4}
Deyao Zhu, Jun Chen, Xiaoqian Shen, Xiang Li, and Mohamed Elhoseiny. 2023.
\newblock Minigpt-4: Enhancing vision-language understanding with advanced large language models.
\newblock \emph{CoRR}, abs/2304.10592.

\end{thebibliography}
\bibliographystyle{acl_natbib}

\newpage
\appendix

\section{Hallucinations in Different Components}

We show the potential hallucinations of each component of LVLMs,  and the corresponding mitigation methods in Table \ref{tab:app1}

\begin{table*}[]
    \centering
    \begin{tabular}{c|c|c}
    \toprule
         Component &  Hallucination? & Mitigation \\ \midrule
         Vision Backbone& \checkmark & w-ECLIP \& w-FineIns
          \\
        Projector& \checkmark & {Int. Align.} \includegraphics[height=1em]{figures/fire_emoji.png} \& {Int. Align.} \includegraphics[height=1em]{figures/snowflake_emoji.png} \& {Sep. Ctrs. Align.}
          \\
          LLM& \XSolidBrush & N/A
          \\ \bottomrule
          
    \end{tabular}
    \caption{Illustration of potential hallucinations in the components of LVLMs, and the corresponding mitigation methods }
    \label{tab:app1}
\end{table*}
\section{More Experiments on Hallucination Benchmark}  \label{app:hallucination_benchmark}
\textcolor{black}{We further add experiments on another hallucination benchmark, Amber. The experimental results of Table \ref{tab:amber_results} show the effectiveness of our method.}

\section{Case Study}
\begin{figure}[h]
    \centering
    \includegraphics[width=\linewidth]{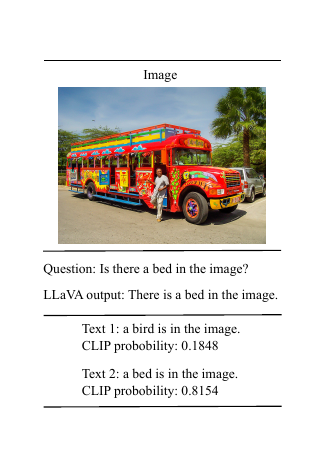}
    \caption{The illustration of the hallucinated case for CLIP and LLaVA.}
    \label{fig:case}
\end{figure}

We showed some hallucinated examples in Figure \ref{fig:case}. 
We can see that the hallucination caused by CLIP can be further input to the LVLM, causing the hallucination in the LVLM.

\section{Comparison with the Existing Hallucination Mitigation Method} \label{app:comparison}
\textcolor{black}{To verify the effectiveness or our methods, we further add more baselines on POPE, as shown in Table \ref{tab:pope_f1_scores}. From this table, our methods show competitive performance with the best baseline (i.e., Less is more). This further demonstrates the effectiveness of our method.}
\begin{table*}[ht]
\centering
\begin{tabular}{lccccccc}
\hline
\textbf{Dataset} & \textbf{LLaVA-7B} & \textbf{w-ECLIP} & \textbf{w-FineIns} & \textbf{Int. Align.} \includegraphics[height=1em]{figures/fire_emoji.png} & \textbf{Int. Align.} \includegraphics[height=1em]{figures/snowflake_emoji.png} & \textbf{Sep. Ctrs. Align.}
\\
\hline
Existence & 83 & \textbf{93} & 92 & 88 & 91 & 87 \\
Attribute & 64 & \textbf{81} & \textbf{81} & 75 & 78 & 76 \\
Relation  & 65 & 69 & \textbf{70} & 57 & 62 & 59 \\
All & 71 & 73 & \textbf{81} & 73 & 77 & 74 \\
\hline
\end{tabular}
\caption{Performance on the Amber dataset across different model variants. Bold indicates best scores per row.}
\label{tab:amber_results}
\end{table*}

\begin{table}[ht]
\centering
\begin{tabular}{lc}
\hline
\textbf{Method} & \textbf{F1 Score} \\
\hline
DoLa~\cite{DBLP:conf/iclr/ChuangXLKGH24} & 80.2 \\
ITT~\cite{DBLP:conf/nips/0002PVPW23} & 83.7 \\
VCD~\cite{DBLP:journals/corr/abs-2311-16922} & 83.2 \\
AGLA~\cite{an2025mitigatingobjecthallucinationslarge} & 84.6 \\
OPERA~\cite{huang2024operaalleviatinghallucinationmultimodal} & 85.2 \\
DOPRA~\cite{DBLP:conf/mm/WeiZ24} & 85.6 \\
HALC~\cite{halc} & 83.9 \\
FastV~\cite{fastv} & 81.3 \\
Less is more~\cite{lessismore} & 86.0 \\
CCA-LLAVA~\cite{DBLP:conf/nips/XingLLL24} & 85.5 \\
LRV \cite{liu2024mitigating} & 80.0\\
Amber \cite{DBLP:journals/corr/abs-2311-07397} & 81.6\\
EAH~\cite{eah} & 85.7 \\ \midrule

w-ECLIP & 85.9 \\
w-FineIns & 85.5 \\
{Int. Align.} \includegraphics[height=1em]{figures/fire_emoji.png} & 85.5 \\
{Int. Align.} \includegraphics[height=1em]{figures/snowflake_emoji.png} & 85.5 \\
Sep. Ctrs. Align & \textbf{86.0} \\
\hline
\end{tabular}
\caption{POPE F1 scores for baselines and proposed methods. Bold indicates the highest score.}
\label{tab:pope_f1_scores}
\end{table}

\section{Experiment on General Benchmark}\label{app:generalbenchmark}
\textcolor{black}{To verify the impact of the proposed method on general capabilities, we further conduct experiments on the general benchmark LLaVA-Bench \cite{llava}. The results of Table \ref{tab:model_comparison_llavabench} show the effectiveness of our method.}
 
\begin{table}[ht]
\centering
\resizebox{\linewidth}{!}{

\begin{tabular}{lcccc}
\hline
\textbf{Model} & \textbf{Conv} & \textbf{Detail} & \textbf{Complex} & \textbf{Full} \\
\hline
LLaVA-7B            & 92 & 75 & 75 & 81 \\ \midrule
w-ECLIP             & 93 & 84 & 87 & 88 \\
w-FineIns           & 94 & 86 & 86 & 89 \\
{Int. Align.} \includegraphics[height=1em]{figures/fire_emoji.png}          & 95 & 87 & 83 & 89 \\
{Int. Align.} \includegraphics[height=1em]{figures/snowflake_emoji.png} & 93 & 84 & 82 & 86 \\
Sep. Ctrs. Align    & 99 & 85 & 87 & 90 \\
\hline
\end{tabular}}
\caption{Model performance comparison on different categories and the full set on LLaVA-Bench.}
\label{tab:model_comparison_llavabench}
\end{table}

\section{Ablation Study} \label{app:ablation}
\textcolor{black}{In this section, we conduct ablation experiments to assess the contribution of each component in the loss function by individually removing the weights $\lambda_1$ and $\lambda_2$. The results are shown in the Table \ref{tab:loss_ablation}. These results demonstrate that both components play meaningful roles in enhancing model performance.}

\begin{table}[ht]
\centering
\resizebox{\linewidth}{!}{
\begin{tabular}{lcccccc}
\hline
\multirow{3}{*}{\textbf{Method}} & \multicolumn{6}{c}{\textbf{POPE}}  \\ \cmidrule(lr){2-7}

&
\multicolumn{2}{c}{\textbf{Random}} & \multicolumn{2}{c}{\textbf{Popular}} & \multicolumn{2}{c}{\textbf{Adversarial}} \\
\cmidrule(lr){2-3} \cmidrule(lr){4-5} \cmidrule(lr){6-7}
& \textbf{Acc} & \textbf{F1} & \textbf{Acc} & \textbf{F1} & \textbf{Acc} & \textbf{F1} \\
\hline
w-ECLIP             & \textbf{87.80} & \textbf{86.87} & \textbf{87.30} & \textbf{86.04} & \textbf{85.87} & \textbf{84.70} \\
$\lambda_1 = 0$     & 87.50 & 86.38 & 86.93 & 85.84 & 85.62 & 83.97 \\
$\lambda_2 = 0$     & 87.52 & 86.47 & 86.79 & 85.88 & 85.47 & 84.11 \\
\hline
\end{tabular}
}
\caption{Ablation study on the impact of loss function components $\lambda_1$ and $\lambda_2$ across different POPE test subsets.}
\label{tab:loss_ablation}
\end{table}

\end{document}